\documentclass{article} 
\usepackage{booktabs}
\usepackage[utf8]{inputenc}
\usepackage[T1]{fontenc}
\usepackage{amsmath,amssymb,amsfonts,amsthm}
\usepackage{nicefrac}
\usepackage{microtype}
\usepackage{bbm}
\usepackage{dsfont}
\usepackage{color}
\usepackage{dsfont}
\usepackage{multirow}
\usepackage{mdframed}
\usepackage{ulem}
\usepackage{times}
\usepackage{pifont}% http://ctan.org/pkg/pifont

\title{Asymptotically Optimal Strategies For Combinatorial Semi-Bandits in Polynomial Time}
\author{Thibaut Cuvelier and Richard Combes and Eric Gourdin}

\newtheorem{assumption}{Assumption}
\newtheorem{theorem}{Theorem}
\newtheorem{lemma}{Lemma}
\newtheorem{proposition}{Proposition}

\makeatletter
\DeclareRobustCommand*\cal{\@fontswitch\relax\mathcal}
\makeatother

\newcommand{\cmark}{\color{green} \ding{51}}%
\newcommand{\xmark}{\color{red} \ding{55}}%

\newcommand{\cI}{\mathcal{I}}
\newcommand{\cJ}{\mathcal{J}}

\newcommand{\cM}{\mathcal{M}}
\newcommand{\cN}{\mathcal{N}}
\newcommand{\cO}{\mathcal{O}}

\newcommand{\cX}{\mathcal{X}}

\newcommand{\EE}{\mathbb{E}}
\newcommand{\NN}{\mathbb{N}}

\newcommand{\RR}{\mathbb{R}}
\newcommand{\indic}{{\bf 1}}

\newcommand{\poly}{\textbf{poly}}
\newcommand{\conv}{\textbf{conv}}

\begin{document}

\maketitle

\begin{abstract}
We consider combinatorial semi-bandits with uncorrelated Gaussian rewards. In this article, we propose the first method, to the best of our knowledge, that enables to compute the solution of the Graves-Lai optimization problem in polynomial time for many combinatorial structures of interest. In turn, this immediately yields the first known approach to implement asymptotically optimal algorithms in polynomial time for combinatorial semi-bandits. 
\end{abstract}

\section{Introduction}

We consider combinatorial bandits, where a learner repeatedly selects decisions $x$ from a combinatorial set $\cX \subset \{0,1\}^d$, and obtains random rewards with mean $\theta^\top x$, where $\theta$ is an unknown vector. The goal of the learner is to maximize the expected sum of rewards. In the semi-bandit setting, the learner can see several individual components of $\theta$ instead of only the total reward $\theta^\top x$.

When decisions in $\cX$ have exactly one nonzero entry, the problem reduces to classical bandits~\cite{lai1985}, for which asymptotically optimal strategies such as KL-UCB and Thompson sampling are known~\cite{lai1985,kaufmann2012,cappe2012}. When $\cX$ is a general set, then the problem reduces to linear bandits~\cite{dani08}.

Combinatorial semi-bandits have been widely studied. Many authors proposed algorithms and regret upper bounds, including Combinatorial Upper Confidence Bound (CUCB)~\cite{kveton2014tight}, Efficient Sampling for Combinatorial Bandits (ESCB)~\cite{combes2015,degenne2016}, Approximate Efficient Sampling for Combinatorial Bandits (AESCB)~\cite{cuvelier2020} and a combinatorial version of Thompson sampling (TS)~\cite{wang2018}. An information-theoretic regret lower bound was also provided by~\cite{combes2015}. Section~\ref{sub_section:algorithms} details algorithms and regret guarantees. While our work focuses on stochastic rewards, the adersarial case was also considered, see \cite{audibert2013} and references therein.

The main reason why the problem is both interesting and difficult is the combinatorial structure of the decision set $\cX$. Many practical problems can be modeled as a combinatorial bandit problem with a particular structure for $\cX$, for instance resource allocation (when $\cX$ is the set of matchings) or network routing (when $\cX$ is a set of source-destination paths in a graph). Typically, $|\cX|$ is exponential in the dimension $d$, so that an exhaustive search over $\cX$ is infeasible in practice, and a major challenge is to derive \emph{computationally efficient} algorithms. Several authors considered particular structures for $\cX$, notably $m$-sets and matroids, as in this case one can derive stronger results~\cite{wen2015,talebi2016,perrault2019}. We will consider more general structures than these two examples. 

Combinatorial semi-bandits are a particular case of structured bandits studied by~\cite{graves1997,combes2017}. It is noted that the Graves-Lai regret lower bound~\cite{graves1997} generalizes to all structured bandits the well-known Lai-Robbins regret lower bound~\cite{lai1985} which holds for classical bandits. For all such problems, there exists \emph{asymptotically optimal algorithms} such as Optimal Sampling for Structured Bandits (OSSB)~\cite{combes2017} under one condition: one must be able to solve a given optimization problem, which we refer to as the Graves-Lai optimization problem. Solving this problem yields both a regret lower bound that holds for any algorithm, as well as an algorithm to attain it, by solving the Graves-Lai problem repeatedly.

Therefore, we believe that one of the most important question to be solved in combinatorial semi-bandits is how to solve the Graves-Lai problem efficiently, in polynomial time in the dimension $d$. This issue is paramount in solving high-dimensional problems. This is far from straightforward, as the number of variables and constraints in the Graves-Lai problem is proportional to $|\cX|$, which is typically exponential in $d$ (see Section~\ref{section:graves_lai_formulation}).

{\bf Our contribution} We propose the first method, to the best of our knowledge, that enables to compute the solution of the Graves-Lai optimization problem in polynomial time for many combinatorial structures of interest. In turn, this immediately yields the first known approach to implement asymptotically optimal algorithms (such as OSSB) in polynomial time for combinatorial semi-bandits. 

The rest of the article is organized as follows. In Section~\ref{section:model}, we define the model, give examples of combinatorial structures of interest, and recall the main algorithms for the problem at hand. In Section~\ref{section:graves_lai_formulation}, we introduce the Graves-Lai optimization problem, and show that solving this problem is both necessary and sufficient to obtain asymptotically optimal algorithms. In Section~\ref{section:main_results}, we propose and analyze Graves-Lai Projected Gradient (GLPG), an algorithm to compute the solution to the Graves-Lai optimization problem in polynomial time. Section~\ref{section:conclusion} concludes the paper. Complete proofs are presented in appendix.
\section{Model}\label{section:model}

\subsection{Combinatorial Semi-Bandits with Uncorrelated Gaussian Rewards}\label{subsection:combinatorial_semi_bandits}

We consider combinatorial semi-bandits with uncorrelated Gaussian rewards. A learner is given a combinatorial set $\cX \subset \{0,1\}^d$, the set of available decisions at each step. Then, for $t=1,...,T$, \textit{(i)} the learner chooses a decision $x(t) \in \cX$, \textit{(ii)} the environment draws $Y(t) \sim \mathcal{N}(\theta,{1 \over 2} I_d)$, \textit{(iii)} the learner observes $x(t) \odot Y(t)$ where $\odot$ represents the Hadamard product\footnote{The Hadamard product of two vectors $x$ and $y$ in $\RR^d$ is the element-wise product: $x \odot y = (x_1 y_1,...,x_d y_d)$}, \textit{(iv)} the learner receives a scalar reward $Y(t)^\top x(t) $. The goal of the learner is to maximize the expected cumulative reward.

The vectors $Y(1),...,Y(t)$ are assumed to be drawn in an i.i.d. fashion from $\mathcal{N}(\theta,{1 \over 2} I_d)$, so that $Y_1(t),...,Y_d(t)$ are uncorrelated Gaussian random variables with respective means $\theta_1,...,\theta_d$ and  variance ${1 \over 2}$. Vector $\theta$ is unknown to the learner, and the chosen decision $x(t)$ only depends on $\cX$ and the history of observations up to time $t$, i.e. $(x(1) \odot Y(1)),...,(x(t-1) \odot Y(t-1))$. In semi-bandit feedback, we observe $x(t) \odot Y(t)$: when $x_i(t) = 1$, we observe $Y_i(t)$, a noisy realization of $\theta_i$ that can be used to estimate $\theta_i$. Conversely, when $x_i(t) = 0$, we do not observe anything. Therefore, in order to maximize the reward, we must be able to get accurate estimates of the initially unknown $\theta$; to do so, we must make sure that $x_i(t) = 1$ often enough to get sufficient statistical information about each $\theta_i$.

\begin{figure}[ht]
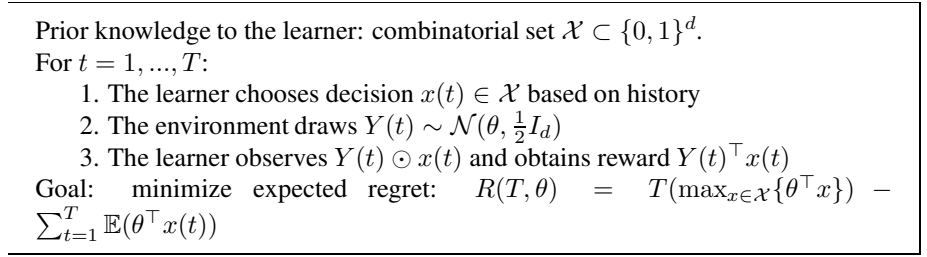

\begin{mdframed}
    Prior knowledge to the learner: combinatorial set ${\cal X} \subset \{0,1\}^d$.

    For $t=1,...,T$:
    
    \hspace{0.5cm} 1. The learner chooses decision $x(t) \in {\cal X}$ based on history

    \hspace{0.5cm} 2. The environment draws $Y(t) \sim \mathcal{N}(\theta,\frac{1}{2} I_d)$

    \hspace{0.5cm} 3. The learner observes $Y(t) \odot x(t)$ and obtains reward $Y(t)^\top x(t)$

    Goal: minimize expected regret: $R(T,\theta) = T (\max_{x \in \cX} \{ \theta^\top x \} ) - \sum_{t=1}^{T}  \EE(\theta^\top x(t))$

\end{mdframed}
    \caption{\label{myfig} Interaction Between Learner and Environment.}
\end{figure}

The goal is to maximize the cumulative reward, or equivalently minimize the total regret. The total regret is defined as the difference in terms of cumulative reward between the learner and that of an oracle who knows $\theta$ in hindsight and always selects $x^\star \in \arg\max_{x \in \cX} \{ \theta^\top x\}$, a decision maximizing the expected reward.
\begin{align*}
    R(T,\theta) = T \left(\max_{x \in \cX} \{ \theta^\top x \} \right) -  \sum_{t=1}^{T}  \EE\left(\theta^\top x(t) \right)
\end{align*}
The model is summarized in Figure~\ref{myfig}. 

We introduce some useful notations. For any decision $x \in \cX$, we denote by $\Delta_{x} = \theta^\top x^\star - \theta^\top x$ the reward gap between $x$ and an optimal decision $x^\star$. We define the minimal $\Delta_{\min} = \min_{x \in \cX: \Delta_x > 0} \Delta_{x}$ and maximal reward gap $\Delta_{\max} = \max_{x \in \cX} \Delta_x$. Furthermore, we define $m = \max_{x \in \cX} \{ \indic^{\top} x\}$ the maximal size of a decision, as measured by the number of non-null entries.

\subsection{Combinatorial Structures}\label{sub_section:combinatorial_structures}

Of course, not much can be achieved if the combinatorial set $\cX$ is arbitrary. For instance, even when $\theta$ is known, if the optimal decision $\max_{x \in \cX} \{ \theta^\top x \}$ cannot be computed efficiently (e.g., $\mathcal{NP}$-hard), the corresponding combinatorial semi-bandit problem is highly unlikely to have an efficient algorithm. We now highlight the combinatorial structures considered here, which include a large amount of classical and important structures for applications to real-world problems. We consider the same combinatorial structures as~\cite{cuvelier2020}. More on combinatorial structures and optimization can be found in~\cite{schrijver-book} and references therein. For any structure that is defined using a graph $G = (V,E)$, by a slight abuse of notation, we identify a subset of edges with the corresponding vector $x = \{0,1\}^{E}$. In that case, the ambient dimension is the number of edges $d = |E|$. Here are the considered combinatorial structures: 

\begin{itemize}\setlength\itemsep{0.01em}
\item \emph{$m$-sets.} Binary vectors with $m$ non-null entries.
\item \emph{Spanning trees.} Spanning trees of a given graph $G = (V,E)$.
\item \emph{Matroids.} Bases of a matroid over a ground set. This includes spanning trees as a particular case.
\item \emph{Source-destination paths.} Paths in a directed, acyclic graph $G = (V,E)$ between a given source and destination.
\item \emph{Matchings.} Matchings in a bipartite graph $G = (V,E)$.
\item \emph{Intersection of two matroids.} Intersection between the sets of bases of two matroids. This includes matchings as a particular case.
\end{itemize}          

\subsection{Optimization Problems}\label{sub_section:optimization_problems}
As shown below, most if not all of the algorithms for combinatorial semi-bandits involve solving some optimization problems over $\cX$. We consider three optimization problems:
\begin{itemize}\setlength\itemsep{0.1em}
\item \emph{Linear Maximization} Compute $\max_{x \in \cX} \{ a^\top x \}$ \hfill $(P_{LM})$
\item \emph{Index Maximization} Compute $\max_{x \in \cX} \{ a^\top x + \sqrt{u^\top x} \}$ \hfill $(P_{IM})$
\item \emph{Budgeted Linear Maximization} Compute $\max_{x \in \cX} \{ a^\top x \}$ subject to $u^\top x \ge s$ \hfill $(P_{BLM})$
\end{itemize}
where $a$, $u$ are vectors with positive integer entries and $s$ is a positive scalar. Table~\ref{tab:complexityoptim} indicates whether an algorithm to solve these problems in polynomial time is known. Approximate $P_{BLM}$ means that we can solve $P_{BLM}$ up to a given approximation ratio. The authors in~\cite{cuvelier2020} provide algorithms for the polynomial cases depicted in this table. Solving $P_{BLM}$, either exactly or approximately, is the cornerstone of our approach to design asymptotically optimal algorithms.

\begin{table}[ht]
    \centering
    \begin{tabular}{c|c|c|c|c}
                 &  $P_{LM}$ & $P_{BLM}$ & approximate $P_{BLM}$ & $P_{IM}$\\ \hline
       $m$-sets  & \cmark   & \cmark  &  \cmark &  \xmark \\ \hline
       spanning trees  & \cmark   & \xmark & \cmark &  \xmark  \\ \hline
       matroids  & \cmark   & \xmark & \cmark &  \xmark \\ \hline
       s-t paths  &  \cmark  & \cmark  &  \cmark &  \xmark \\ \hline
       %knapsack sets &   &  &  \\   \hline
       matchings  &  \cmark  & \xmark & \cmark &  \xmark \\ 
    \end{tabular}
    \caption{Polynomial Solvability of Combinatorial Problems over $\cX$.}
    \label{tab:complexityoptim}
\end{table}

\subsection{Algorithms, Regret, and Complexity}\label{sub_section:algorithms}

To understand the interplay between regret and computational efficiency, we now describe the most studied algorithms for combinatorial semi-bandits and highlight their regret guarantees. We define the number of samples obtained up to time $t$:
$$
    n_i(t) = \sum_{t'=1}^{t-1} x_i(t) ,\quad i=1,...,d
$$
as well as the corresponding empirical mean reward at time $t$:
$$
     \hat\theta_i(t) = {1 \over \max(1,n_i(t)) } \sum_{t'=1}^{t-1} x_i(t) Y_i(t) ,\quad i=1,...,d
$$
The simplest algorithm is CUCB~\cite{kveton2014tight}, an extension of the well-known UCB~\cite{auer_finitetimeanalysis_2002} algorithm for stochastic bandits. ESCB~\cite{combes2015} is an improved version of CUCB taking advantage of the fact that rewards are not correlated. AESCB~\cite{cuvelier2020} is an approximate version of ESCB with lower computational complexity. TS~\cite{wang2018} is an algorithm inspired by Bayesian approaches. OSSB~\cite{combes2017} is a general, asymptotically optimal algorithm designed for general structured bandits and that can be specialized to combinatorial semi-bandits. These algorithms select a decision $x(t)$ according to the following rules.
\begin{itemize} 
\item \emph{CUCB:} $x(t) \in \arg\max_{x \in \cX} \{ \hat\theta(t)^\top x + \sum_{i=1}^d x_i {\ln T \over n_i(t)} \}$
\item \emph{ESCB:} $x(t) \in \arg\max_{x \in \cX} \{ \hat\theta(t)^\top x + \sqrt{\sum_{i=1}^d x_i {\ln T \over n_i(t)}} \}$
\item \emph{AESCB:} $x(t)$ with $\max_{x \in \cX} \{ \hat\theta(t)^\top x + \sqrt{\sum_{i=1}^d x_i {\ln T \over n_i(t)}} \} \le  \hat\theta(t)^\top x(t)$  \\ $+{1 \over \varepsilon_t} \sqrt{\sum_{i=1}^d x_i(t) {\ln T \over n_i(t)}}$, with $\delta_t,\varepsilon_t$ two input parameters.
\item \emph{TS:} $x(t) \in \arg\max_{x \in \cX} \{ V(t)^\top x \}$ where $V(t) \sim \cN(\hat\theta(t), {\bf diag}({1 \over n_1(t)},...,{1 \over n_d(t)})) $ is a sample from the posterior distribution of $\theta$ given the information available at time $t$
\end{itemize}          

Table~\ref{tab:algorithms_regret_complexity} summarizes the regret and complexity of algorithms. OSSB is provably asymptotically optimal, while ESCB and AESCB enjoy a $\cO( {d (\ln m)^2 \over \Delta_{min}}  \ln T  )$ regret guarantee. TS has a larger regret guarantee of $\cO( {d \sqrt{m} \over \Delta_{min}}  \ln T)$, and CUCB has the largest one: $\cO( {d m \over \Delta_{min}}  \ln T  )$.  There is an interesting interplay here between statistical efficiency (regret) and computational complexity. In terms of complexity, for each time step, CUCB and TS involve solving $P_{LM}$, while ESCB involves solving $P_{IM}$, which typically cannot be solved in polynomial time in the dimension $d$, and AESCB involves solving (up to a fixed approximation ratio) $P_{BLM}$ several times, which can be done in polynomial time. Finally, OSSB involves solving $P_{GL}$, see section~\ref{section:graves_lai_formulation}. 

\begin{table}[ht]
    \centering
    \begin{tabular}{c|c|c|c}
                 &  Regret & Complexity & Asymptotically Optimal \\ \hline
       CUCB  & $O( {d m \over \Delta_{\min}}  \ln T  )$ & Solve $P_{LM}$ once &  \xmark \\ \hline
       TS  & $O( {d \sqrt{m} \over \Delta_{\min}}  \ln T)$ & Solve $P_{LM}$ once &  \xmark \\ \hline
       ESCB  & $O( {d (\ln m)^2 \over \Delta_{\min}}  \ln T  )$ & Solve $P_{IM}$ once &  \xmark \\ \hline
       AESCB  & $O( {d (\ln m)^2 \over \Delta_{min}}  \ln T  )$ & Approximate $P_{BLM}$  several times &  \xmark \\ \hline
       OSSB & $O( C(\theta) \ln T  )$ & Solve $P_{GL}$ once (see section \ref{section:graves_lai_formulation}) & \cmark  \\
    \end{tabular}
    \caption{Algorithms, Regret and Complexity}
    \label{tab:algorithms_regret_complexity}
\end{table}

\section{Graves-Lai Formulation: Regret Lower Bound and Asymptotically Optimal Algorithms}\label{section:graves_lai_formulation}

Combinatorial semi-bandits are an instance of structured bandits studied in~\cite{combes2017}, which in turn are an instance of the controlled Markov chains studied by the seminal work of~\cite{graves1997}. Those results can be applied to our problem, and in fact doing so yields asymptotically optimal algorithms, i.e. whose regret is, asymptotically, the lowest achievable. We explain how these algorithms function, and explain the challenge of implementing these algorithms efficiently. 

\subsection{The Graves Lai Optimization Problem}\label{subsection:graves_lai}

We first introduce the following optimization problem, which we call the Graves-Lai optimization problem for combinatorial semi-bandits:
\begin{align*}
    \hspace{3cm} \underset{\alpha \in \RR^{|\cX|}_+}{\text{minimize   }} & \sum_{x \in \cX} \alpha_x \Delta_x \hspace{4cm} \quad (P_{GL}) \\
    \hspace{3cm} \text{subject to    } & \sum_{i \in \cI} {x_i \over \sum_{y \in \cX} y_i \alpha_y} \le \Delta_x^2 \;,\; \forall x \in \cX 
\end{align*}
with 
\begin{align*}
    \cI = \left\{i \in \{1,...,d\}: \max_{x \in \cX:x_i=1} (\theta^\top x) < \max_{x \in \cX} (\theta^\top x) \right\}
\end{align*}
the set of items $i$ that do not appear in any optimal decision. This type of problem first appeared in~\cite{graves1997} in the more general context of controlled Markov chains, and was later specialized to combinatorial semi-bandits by~\cite{combes2015}. Not only does the Graves-Lai optimization problem yield a regret lower bound that holds for any algorithm, but computing its solution also enables to design algorithms achieving this bound and are hence asymptotically optimal~\cite{combes2015}.

\subsection{Regret Lower Bound}\label{subsection:regret_lower_bound}

Theorem~\ref{th:lower_bound} states that the regret of any uniformly good algorithm (i.e. an algorithm whose regret scales as $o(T^{a})$ when $T \to \infty$ for any fixed problem instance and any $a > 0$) is lower bounded by the optimal value of the Graves-Lai optimization problem. The proof follows from~\cite{combes2015}[Theorem 1] and is presented in appendix.

\begin{theorem}[\cite{graves1997,combes2015}]\label{th:lower_bound}
    Consider a uniformly good algorithm, in the sense that its expected regret verifies $R(T,\theta) = o(T^a)$ for any fixed $\theta \in \RR^d$ and $a > 0$.
    
    Then, its regret verifies for any $\theta$:
    $$
        \lim\inf_{T \to \infty} {R(T,\theta) \over \ln T} \ge C(\theta)
    $$
    where $C(\theta)$ is the optimal value of the Graves-Lai optimization problem  $P_{GL}$.
\end{theorem}

The analysis of~\cite{graves1997} provides the following interpretation of the objective function and the constraints in $P_{GL}$. Consider a uniformly good algorithm selecting each sub-optimal decision $x \in \cX$ an amount of time equal to $\alpha_x \ln T$. The regret of this algorithm is $(\ln T) \sum_{x \in \cX} \alpha_x \Delta_x$, which is proportional to the objective function of $P_{GL}$. The number of observations to estimate $\theta_i$ equals $(\ln T) \sum_{y \in \cX} y_i \alpha_{y}$. Given a sub-optimal decision $x$, in order to be sure that $x \ne x^\star$, one needs enough statistical information to estimate $\theta_i$ for all $i$ such that $x_i = 1$ and $x^\star_i = 0$. To the contrary, if $x^\star_i = 1$, then $\theta_i$ can be estimated very accurately without regret, as sampling decision $x^\star$ does not incur regret. More precisely, one can show that the number of observations of any $x\in\cX$ must satisfy
$$
    \sum_{i \in \cI} {x_i \over \sum_{y \in \cX} y_i \alpha_y} \le \Delta_x^2
$$
Otherwise, it is impossible to distinguish between decision $x$ and the optimal decision $x^\star$ with high probability. In short, the Graves-Lai optimization problem $P_{GL}$ simply consists in minimizing regret, subject to the constraint that one can statistically distinguish between optimal and sub-optimal decisions.

\subsection{Asymptotically Optimal Algorithms}\label{subsection:asymptotically_optimal_algorithms}

In fact, if one can compute the solution of $P_{GL}$, there exists asymptotically optimal algorithms attaining the lower bound of Theorem~\ref{th:lower_bound} such as the doubling trick algorithm of~\cite{graves1997} and the arguably simpler OSSB algorithm from~\cite{combes2017}. Both of these algorithms are based on \emph{certainty equivalence}, which involves estimating $\theta$ using empirical averages, and selecting each sub-optimal decision an amount of time $\alpha_x^\star \ln T$, where $\alpha^\star$ is an optimal solution of $P_{GL}$, and where $\theta$ is replaced by its estimate. Therefore, the solution of the Graves-Lai optimization problem explicitly gives the way that one should explore sub-optimal decisions to minimize regret.

\begin{theorem}[\cite{graves1997,combes2017}]\label{th:upper_bound}
    Assume that one can compute solutions to the Graves-Lai optimization problem for any $\theta$. Then, there exists asymptotically optimal algorithms in the sense that their regret verifies for all $\theta$:
    $$
        \lim\sup_{T \to \infty} {R(T,\theta) \over \ln T} \le C(\theta)
    $$
\end{theorem}    
    
\subsection{Computational Complexity of Asymptotically Optimal Algorithms}\label{subsection:asymptotically_optimal_algorithms_complexity}

We can conclude that the only difficulty in the design of asymptotically optimal algorithms is a \emph{computational} one. One must be able to compute solutions to the Graves-Lai optimization problem $P_{GL}$ efficiently. At first look, this seems like a difficult task. Namely, $P_{GL}$ involves optimizing a linear function with $|\cX|$ variables, subject to $|\cX| + d$ convex constraints. Indeed, for any $x \in \cX$, the function
$$
\alpha \mapsto \sum_{i \in \cI} {x_i \over \sum_{y \in \cX} y_i \alpha_y}
$$
is convex.  Therefore, simply checking whether or not some solution $\alpha $ is feasible may require $\cO(|\cX|)$ computations, and $\cO(|\cX|)$ is not polynomial in $d$ for any of the combinatorial structures considered in Section~\ref{sub_section:combinatorial_structures}. Furthermore, even assuming that the optimal solution $\alpha^\star$ can be computed, if the size of $\{x \in \cX: \alpha_x^\star  > 0 \}$ is close to that of $\cX$, then simply outputting the optimal solution is not possible in polynomial time. Our main result demonstrates that it is indeed possible to solve $P_{GL}$ in polynomial time, as shown in the next section.

\section{Main Result}\label{section:main_results}

\subsection{Assumptions}\label{subsection:assumptions}

Before stating our results, we discuss some of our assumptions.

\begin{assumption}[Covering]\label{assumption:covering}
    For each $i \in \{1,...,d\}$, there exists a decision $x^i \in \cX$ such that $x_i^i = 1$.
\end{assumption}

Assumption~\ref{assumption:covering} states that, for all $i$, there must exist a decision $x^i \in \cX$ with $x_i^i = 1$, so that $\theta_i$ may be estimated by sampling $x^i$. If this assumption does not hold, we can simply remove $i$ from consideration, since it plays no role in the Graves-Lai optimization problem. Thus, this assumption can be made without loss of generality.

\begin{assumption}[Integrality]\label{assumption:integrality}
    We have that $\theta \in \NN^d$.
\end{assumption}
Assumption~\ref{assumption:integrality} states that the vector $\theta$ has positive integer entries. While this makes stating our results simpler, we can easily generalize them to the case where $\theta$ has continuous values. Proposition~\ref{prop:discretization} in appendix states that if $\theta$ is real valued, we can discretize $\theta$ as 
$
    \theta^{\varepsilon} = \varepsilon( \lceil \theta_1/\varepsilon,\rceil,...,\lceil \theta_d/\varepsilon,\rceil    )
$
then solve an approximate version of $P_{GL}$ where $\theta$ is replaced by $\theta^{\varepsilon}/\varepsilon$, which has integer entries. This enables us to solve $P_{GL}$ up to an error of $O(1/\varepsilon)$ in time $O\Big(\poly(d,1/\varepsilon)\Big)$. Hence, one can solve $P_{GL}$ up to any fixed accuracy in polynomial time using our results.

\begin{assumption}[Polynomial-Time Linear Maximization]\label{assumption:polynomial_LM}
    The exact solution of $P_{LM}$ can be computed in time $O\Big(\poly(d)\Big)$. 
\end{assumption}
\begin{assumption}[Polynomial-Time Budgeted Linear Maximization]\label{assumption:polynomial_BLM}
        The exact solution of $P_{BLM}$ can be computed in time $O\Big(\poly(d,\|u \|_\infty)\Big)$. 
\end{assumption}
\begin{assumption}[Polynomial-Time Approximate Budgeted Linear Maximization]\label{assumption:polynomial_BLM_approximate}
    An $\varepsilon$-optimal solution of $P_{BLM}$ can be computed in time $O\Big(\poly(d,\|u \|_\infty)\Big)$ for some fixed $\varepsilon > 0$, in the sense that we can compute $\tilde{x} \in \cX$ verifying:
    $$
    a^\top \tilde{x} \ge \varepsilon \left( \max_{x \in \cX: u^\top x \ge s} \{ a^\top x \} \right) \text{ and } u^\top \tilde{x} \ge s
    $$
\end{assumption}

Assumptions~\ref{assumption:polynomial_LM}, \ref{assumption:polynomial_BLM}, and \ref{assumption:polynomial_BLM_approximate} respectively state that one can solve $P_{LM}$ exactly, $P_{BLM}$ exactly, and $P_{BLM}$ approximately. The cases in which those assumptions hold are reported in Table~\ref{tab:complexityoptim}. In particular, in all considered combinatorial structures, Assumption~\ref{assumption:polynomial_BLM_approximate} does hold, as shown in \cite{cuvelier2020}.

\begin{assumption}[Compact Representation for Convex Hulls]\label{assumption:convex_hull}
    The convex hull of $\cX$ can be written in the following form:
    $$
        \conv(\cX) = \{w \in \RR^d: A w = b , w \ge 0\}
    $$
    where the size of $A$ and $b$ is polynomial in the dimension $d$.
\end{assumption}
Assumption~\ref{assumption:convex_hull} states that the convex hull of $\cX$, a polytope, can be represented in a ``compact'' manner, i.e. using a polynomial number of linear inequalities. This assumption is verified for all considered combinatorial structures listed above: spanning trees, matchings, paths, etc. (see for instance \cite{schrijver-book}).

\subsection{Main Result}\label{subsection:main_results}

Our main result is Theorem~\ref{th:main_result}. It states that the solution to the Graves-Lai optimization problem can be computed in polynomial time up to any given accuracy. To do so, we design the GLPG (Graves-Lai Projected Gradient) algorithm, which is presented and analyzed below. More precisely, the complexity of GLPG is polynomial in the dimension $d$, the accuracy level $\delta$ and the largest entry in $\theta$, denoted by 
$\|\theta\|_\infty$. The pseudo-code for GLPG is presented in Figure~\ref{algo_complete}.

Our main result comes in two versions: \textit{(i)} an exact version where one can compute the exact solution up to any given accuracy, when exact Polynomial Time Budgeted Linear Maximization is possible, and \textit{(ii)} an approximate version where one can compute a solution with a fixed approximation ratio up to any given accuracy, when approximate Polynomial Time Budgeted Linear Maximization is possible. If one can only solve the Graves-Lai optimization problem with a fixed approximation ratio, the yielded algorithm is not asymptotically optimal. However, the asymptotic regret of such an algorithm is upper bounded by a universal constant times the Graves-Lai lower bound, which is typically better than what existing algorithms can achieve for large time horizons.

\begin{theorem}\label{th:main_result}
    Consider $\delta > 0$. Let Assumptions~\ref{assumption:covering},  \ref{assumption:integrality}, \ref{assumption:polynomial_LM} and \ref{assumption:convex_hull} hold. 

    (\emph{Exact version}) If Assumption~\ref{assumption:polynomial_BLM} further holds, then the GLPG algorithm outputs $\alpha$, an $\delta$-optimal solution to $P_{GL}$ in time $\poly(d,\delta,\|\theta\|_{\infty})$ in the sense that:
\begin{align*}
\sum_{x \in \cX} \alpha_x \Delta_x \le C(\theta) + \delta \text{ and }
 \sum_{i \in \cI} {x_i \over \sum_{y \in \cX} y_i \alpha_y} \le \Delta_x^2 \;\; \forall x \in \cX,\; \alpha_x \ge 0 \;\; \forall x \in \cX
\end{align*}

    (\emph{Approximate version}) If Assumption~\ref{assumption:polynomial_BLM_approximate} further holds, then the GLPG algorithm outputs $\alpha$, an $(\varepsilon,\delta)$-optimal solution to $P_{GL}$ in time $\poly(d,\delta,\|\theta\|_{\infty})$ in the sense that:
\begin{align*}
\sum_{x \in \cX} \alpha_x \Delta_x \le {1 \over \varepsilon} C(\theta) + \delta \text{ and }
 \sum_{i \in \cI} {x_i \over \sum_{y \in \cX} y_i \alpha_y} \le \Delta_x^2 \;\; \forall x \in \cX,\; \alpha_x \ge 0 \;\; \forall x \in \cX
\end{align*}
\end{theorem}

The main steps of the proof are highlighted in the next subsections. We solely prove the approximate version, as the exact version is a particular case of the approximate one with $\varepsilon = 1$.

\subsection{Step 0: Computing the set of optimal items}\label{subsection:optimal_set}

It is noted that $\cal I$ can be computed in polynomial time using a penalty method. Indeed, one can readily check that $i \in \cI$ if and only if 
$$
     \max_{x \in \cX} \{ \theta^\top y^i\} < \max_{x \in \cX} \{ \theta^\top x\} \text{ where } y^i \in \arg\max_{x \in \cX} \{ (\theta + e^i 2 d \| \theta \|_{\infty})^\top x\} \text{ and } e^i_j = \indic\{i = j\}
$$
From assumption~\ref{assumption:polynomial_LM}, this computation can be done in polynomial time.

\subsection{Step 1: Dimensionality Reduction}\label{subsection:dimensionality_reduction}

The first step in the proof is Proposition~\ref{proposition:reduction} proven in appendix. This proposition shows that the solution of $P_{GL}$, a problem with $|\cX|$ variables, can be derived by computing the solution of $P_{GL}^{'}$, another, much simpler optimization problem with only $d$ variables. The idea behind this reduction is that, instead of optimizing over $(\alpha_x)_{x \in \cX}$ (the amount of time each decision is selected), we can optimize over $(\sum_{x \in \cX} x_i \alpha_x)_{i=1,...,d}$ (the amount of samples obtained to estimate $\theta_1,...,\theta_d$). 

\begin{proposition}\label{proposition:reduction}
   Consider $w^\star \in \RR^d$ the optimal solution to 
   \begin{align*}
    \hspace{3cm} \underset{w \in \RR^{d}}{\text{minimize   }} &  q^\top w  \hfill \hspace{6cm} (P_{GL}') \\
    \text{subject to } & \sum_{i \in \cI} {x_i \over w_i} \le \Delta_x^2 \;\; \forall x \in \cX , \;\;  M w = 0, w \ge 0,  \min_{i \in \cI} w_i \ge \underline{w} 
\end{align*}
    where $M \equiv A - {b b^\top A \over \|b\|^2}$ and $q \equiv (\theta^\top x^\star) {b^\top A \over \|b\|^2} - \theta$ 
and $\underline{w} \equiv (m \| \theta \|_{\infty})^{-2}$. 

Then there exists $\alpha^\star \in \RR^{|\cX|}$ an optimal solution to $P_{GL}$ such that: $w^\star = \sum_{x \in \cX} x \alpha^\star_x$.
\end{proposition}

\subsection{Step 2: Approximate Subgradient Descent}\label{subsection:approximate_gradient_descent}

The next step is to solve the reduced form $P_{GL}'$ using an iterative scheme. To do so, we use a combination of penalization as well as projected subgradient descent. For $x \in \cX$, define
$$
h_x(w) = \left(\sum_{i \in \cI} {x_i \over w_i}\right) - \Delta_{x}^2
$$
as the constraint attached to $x$ in $P_{GL}'$. Instead of solving $P_{GL}'$, we solve $P_{GL}''$ in which the constraints are replaced by a penalty, with $\lambda > 0$:  
\begin{align*}
    \hspace{3cm} \underset{w \in \RR^{d}}{\text{minimize   }} &  \left\{ q^\top w + \lambda \max_{x \in \cX} \Big( h_x(w)\Big)^{+} \right\}  \quad \quad  \hspace{1cm} (P_{GL}'')\\
    \text{subject to } & M w = 0, w \ge 0, \min_{i \in \cI} w_i \ge \underline{w}.
\end{align*}
where $(\cdot)^+ = \max(\cdot,0)$ denotes the positive part. The value of $\lambda$ must be appropriately large to ensure that the constraints in $P_{GL}'$ are satisfied; it will be specified later. Define the polytope
$$
\cM = \{w : M w = 0, w \ge 0, \min_{i \in \cI} w_i \ge \underline{w}  \}
$$
We solve  $P_{GL}''$ using a strategy that resembles the projected subgradient method. The method is iterative with $T$ iterations \footnote{In this section $t$ and $T$ denote the iteration number and the total number of iterations of our method. They should not be confused with $t$ and $T$ as defined in the previous sections.} and follows the update rule for $t=1,...,T$ :
\begin{align*}
	w^0 &= (\underline{w},...,\underline{w}) \\
	w^{t+1} &= \Pi_{\cM} \Big\{ w^{t} - \eta g^t \Big\}. \\
	g^t &=  q + \lambda \varepsilon \nabla h_{x^t}(\varepsilon w^t) \indic( h_{x^t}(\varepsilon w^t) > 0) \\
	\bar{w} &= {1 \over T} \sum_{t=1}^{T} w_t. 
\end{align*}
where $x^t$ is chosen such that
$$
    \max_{x \in \cX} h_x(w^t) \le h_{x_t}(\varepsilon w^t)
$$
and $\Pi_{\cM}$ denotes the orthogonal projection on $\cM$. The output of the algorithm is the average iterate $\bar{w}$ instead of the last iterate $w^{T}$. For some combinatorial sets $\cX$, the projection step can be computed exactly in polynomial time; otherwise, it can be computed using an interior point method, a very efficient method for convex optimization programs (see Section~\ref{ref:projection_step} for more details). In particular, when $\varepsilon = 1$, we have $x^t \in \arg\max_{x\in\cX} h_x(w^t)$, so that $g_t$ is simply a subgradient of $w \mapsto \{ q^\top w + \lambda \max_{x \in \cX} \Big( h_x(w) \Big)^{+}\}$ evaluated at $w^t$ and the proposed algorithm follows projected sub-gradient descent for this function. When $\varepsilon < 1$, our algorithm guarantees that, for any $x$, $h_x(\varepsilon w^t)$ cannot become too large.
 
Furthermore, Proposition~\ref{prop:compute_xt} shows that $x^t$ can be computed in polynomial time under our assumptions, by solving $P_{BLM}$ (exactly or approximately) a polynomial number of times. The proof is in appendix.

\begin{proposition}\label{prop:compute_xt}
    Under Assumption~\ref{assumption:integrality}, and either Assumption~\ref{assumption:polynomial_BLM} or \ref{assumption:polynomial_BLM_approximate}, $x^t$ can be computed in time $\poly(d,\delta,\|\theta\|_{\infty})$.
\end{proposition} 

Proposition~\ref{prop:gradient_descent} states that, when $\lambda$, $\eta$, and $T$ are chosen appropriately, this procedure outputs a solution arbitrarily close to the optimal solution of $P_{GL}'$. Further, this procedure runs in polynomial time. The proof is involved and is given in appendix.

\begin{proposition}\label{prop:gradient_descent}
    Consider any fixed $\delta > 0$. Let
    \begin{align*}
    \delta_{2} &= {\delta \varepsilon \over m^2 d \| \theta \|_{\infty}} \\
    \delta_{1} &= {\delta \over 2 (1+\delta_2)} \\
    \lambda &= {1 \over \delta_2} (\delta_1 + {m^2 d \| \theta \|_{\infty} }) \\
        T &= {1 \over \delta_1^2} \varepsilon^{-2} m^{5} d^2 \| \theta \|_{\infty}^2 \left( \| q \|^2 + \lambda^2 \varepsilon^{-2} d m^8 \|\theta\|_{\infty}^ 8\right) \\
	\eta^2 &= { \varepsilon^{-2} m^{5} d^2 \| \theta \|_{\infty}^2 \over T ( \| q \|^2 + \lambda^2 \varepsilon^{-2} d m^8 \|\theta\|_{\infty}^8)}
    \end{align*}
    Let $\bar{w}$ denote the output of the above procedure, and let $\bar{w}' = (1+\delta_2) \bar{w}$.
    
    Then $\bar{w}'$ is an ($\varepsilon$, $\delta$)-optimal solution to optimization problem $P_{GL}^{'}$ in the sense that
  \begin{align*}
        q^\top \bar{w}' &\le q^\top (w^\star/\varepsilon) + \delta \\
        M \bar{w}' &= 0 , \bar{w}' \ge 0, \min_{i \in \cI} \bar{w}_i' \ge \underline{w} \;,\; \sum_{i \in \cI} {x_i \over \bar{w}'_i} \le \Delta_x^2  \,,\, x \in \cX, \; 
    \end{align*}
    and this procedure runs in time $\poly(d,\delta,\|\theta\|_{\infty})$
\end{proposition}

\subsection{Step 3: Retrieving the Solution to the Original Problem}\label{subsection:proposed_algorithm}

Assume that we have computed $w^\star \in \RR^d$, the optimal solution to $P_{GL}'$. We now need to retrieve $\alpha^* \in \RR^{|\cal X|}$, the optimal solution to the original problem $P_{GL}$. Since $\alpha^*$ has $|\cal X|$ entries, and $|\cal X|$ is typically not polynomial in the dimension $d$, this seems like an impossible task. However, we can choose $\alpha^*$ such that most of its entries are zero: the optimum solution is generally not unique. From Carathéodory's theorem, any point in the convex hull of $|\cal X|$ can be written as a convex combination of at most $d+1$ elements of $\cX$. 

We provide an iterative procedure to compute $\alpha^*$ knowing $w^\star$ and analyze it in Proposition~\ref{prop:caratheodory}. We let $\bar{w}^1 = w^\star$ and for $k=1,...,d$: if $\bar{w}^k = 0$, we let $\alpha_{x^k} = 0$ and $x^k \in \cX$ chosen arbitrarily; otherwise, we let $\alpha_{x^k} = \min_{i:\bar{w}_i^k > 0} \bar{w}_i^k$ and $\bar{w}^{k+1} = \bar{w}^{k} - \alpha_{x^k} x^k$ where 
$$x^k \in \arg\min_{x \in \cX} \Big\{ \sum_{i=1}^{d} x_i  \indic\{\bar{w}^k_i > 0 \}\Big\}$$
The output of this procedure is $\alpha_{x^1},...,\alpha_{x^d}$ and $x^1,...,x^d$, which is a decomposition of $w^\star$ as a linear combination with positive coefficients with at most $d$ elements from $\cX$. The cornerstone of this procedure is the fact that if $\bar{w} = \sum_{x \in \cX} x \alpha_x$ with $\alpha \ge 0$ then for any $x$ such that $\alpha_x > 0$ we have that $x_i = 1$ implies $\bar{w}_i > 0$. This concludes the proof of Theorem~\ref{th:main_result}.

\begin{proposition}\label{prop:caratheodory}
The above procedure is such that $w^\star = \sum_{k=1}^d x^k \alpha_{x^k}$
with $\alpha_{x^1},...,\alpha_{x^d}$ positive numbers and runs in time $\poly(d,\delta,\|\theta\|_{\infty})$.
\end{proposition}

\begin{figure}[ht]
\begin{mdframed}
    {\bf Inputs}: $A$ and $b$ (representation for the convex hull of $\cX$), $\theta$ (mean reward vector), $\delta$ (accuracy level), $\varepsilon$ (approximation ratio) 
    
    Parameter choice: set 
    \begin{align*}
    \delta_{2} &= {\delta \varepsilon \over m^2 d \| \theta \|_{\infty}} \\
    \delta_{1} &= {\delta \over 2 (1+\delta_2)} \\
    \lambda &= {1 \over \delta_2} (\delta_1 + {m^2 d \| \theta \|_{\infty} }) \\
        T &= {1 \over \delta_1^2} \varepsilon^{-2} m^{5} d^2 \| \theta \|_{\infty}^2 \left( \| q \|^2 + \lambda^2 \varepsilon^{-2} d m^8 \|\theta\|_{\infty}^8\right) \\
	\eta^2 &= { \varepsilon^{-2} m^{5} d^2 \| \theta \|_{\infty}^2 \over T ( \| q \|^2 + \lambda^2 \varepsilon^{-2} d m^8 \|\theta\|_{\infty}^8)}
        \end{align*}
    
    \vspace{0.2cm} 
    \emph{Step 1}: Dimensionality Reduction \vspace{0.2cm} 
    
    Compute $M = A - {b b^\top A \over \|b\|^2}$ and $
    q = (\theta^\top x^\star) {b^\top A \over \|b\|^2} + \theta$ and $\underline{w} = ( m \| \theta \|_{\infty})^{-2}$
    
    \vspace{0.2cm} 
    \emph{Step 2}: Approximate Gradient Descent \vspace{0.2cm} 
    
    Set $w^0 = (\underline{w},...,\underline{w})$
    
    For $t=1,...,T$:
    
    \hspace{0.5cm} Find $x^t$ such that $\max_{x \in \cX} h_x(w^t)
    \le h_{x_t}(\varepsilon w^t)$
    
    \hspace{0.5cm}  Compute $g^t =  q + \lambda \varepsilon \nabla h_{x^t}(\varepsilon w^t) \indic( h_{x^t}(\varepsilon w^t) > 0)$
    
    \hspace{0.5cm} Update $w^{t+1} = \Pi_{\cM} \Big\{ w^{t} - \eta g^t \Big\}$
    
    Compute  $\bar{w} = {1 \over T} \sum_{t=1}^{T} w_t$ and $\bar{w}' = (1+\delta_2) \bar{w}$
    
    \vspace{0.2cm} 
    \emph{Step 3}: Retrieving the Solution to the Original Problem \vspace{0.3cm} 

    Set $\bar{w}^1 = \bar{w}'$
    
    For $k=1,...,d$:

    \hspace{0.5cm}  Find $x^{k} \in \arg\min_{x \in \cX} \sum_{i=1}^{d} x_i  \indic\{\bar{w}^k_i > 0 \}$

    \hspace{0.5cm} If $\bar{w}^k > 0$ let $\alpha_{x^k} = \min_{i:{\bar{w}_i^k} > 0} \bar{w}_i^k$, otherwise let $\alpha_{x^k} = 0$.

    \hspace{0.5cm} Update $\bar{w}^{k+1} = \bar{w}^{k} -   \alpha_{x^k} x^k$. 

    {\bf Output}: A $(\delta,\varepsilon)$-optimal solution to the Graves Lai optimization problem $\alpha_{x^1},...,\alpha_{x^d}$ and $x^{1},...,x^{d}$.

\end{mdframed}
    \caption{\label{algo_complete} The GLPG Algorithm: Computing the Solution to $P_{GL}$ in Polynomial Time.}
\end{figure}

\section{Conclusion}\label{section:conclusion}

We have proposed the first method, to the best of our knowledge, to compute the solution of the Graves-Lai optimization problem for combinatorial semi-bandits in polynomial time, which in turn allows to implement asymptotically optimal algorithms (such as OSSB) for this problem. Our results hold for a large number of combinatorial structures including $m$-sets, spanning trees, paths, and matchings. We believe that our results shed some light on the trade-off between statistical efficiency and computational complexity in bandit optimization.

\bibliography{references}
\bibliographystyle{plain}

\newpage
\section{Additional Result}\label{section:additional_results}
To avoid confusion, in this section, for any parameter $\lambda \in \RR^d$, we use the notation
$$
\Delta_x(\lambda) = \max_{y \in \cX} (\lambda^\top y) - \lambda^\top x
$$
to denote the reward gap of decision $x \in \cX$ under parameter $\lambda$.
\begin{proposition}\label{prop:discretization}
    Consider $\varepsilon > 0$ and a real valued vector $\theta = \RR^d$. Define the discretized vector
$$\theta^{\varepsilon} = \varepsilon( \lceil \theta_1/\varepsilon,\rceil,...,\lceil \theta_d/\varepsilon,\rceil )
$$
Consider the following optimization problem which approximates $P_{GL}$:
\begin{align*}
    \hspace{3cm} \underset{\alpha \in \RR^{|\cX|}_+}{\text{minimize   }} & \sum_{x \in \cX^\star} \alpha_x \Delta_x(\theta^\varepsilon) \hspace{3cm} \quad (P_{GL}^\varepsilon) \\
    \hspace{3cm} \text{subject to    } & \sum_{i \in \cI} {x_i \over \sum_{y \in \cX} y_i \alpha_y} \le (\Delta_x(\theta^{\varepsilon}))^2 \;,\; \forall x \in \cX^\star
\end{align*}
with 
\begin{align*}
    \cI &= \left\{i \in \{1,...,d\}: \max_{x \in \cX:x_i=1} (\theta^\top x) < \max_{x \in \cX} (\theta^\top x)\right\} \\
    \cX^\star &= \{x \in \cX: \Delta_x(\theta) > 0\} 
\end{align*}
Denote by $\alpha^{\star,\varepsilon}$ an optimal solution to $P_{GL}^{\varepsilon}$ and $\alpha^\star$ an optimal solution to $P_{GL}$. Assume that $\varepsilon \le {\Delta_{\min} \over 2}$. Then $\alpha^{\star,\varepsilon}(1 + {2m \varepsilon \over \Delta_{\min}})^2$ is a feasible solution to $P_{GL}$, and it is near optimal in the sense that: 
$$
    \sum_{x \in \cX^\star} \alpha^{\star,\varepsilon}_x \Delta_x(\theta) \le (1 + {4m \varepsilon \over \Delta_{\min}})^4 \sum_{x \in \cX^\star} \alpha_x^\star \Delta_x(\theta) 
$$
\end{proposition}
{\bf Proof:} We first upper bound the gap differences. For any $x$ we have
\begin{align*}
    |\Delta_{x}(\theta) - \Delta_{x}(\theta^\varepsilon)| &\le    | \max_{y \in \cX} (\theta^\top y)  - \max_{y \in \cX}( (\theta^\varepsilon)^\top y) |  + | \theta^\top x -(\theta^\varepsilon)^\top x | \\ 
    &\le \max_{y \in \cX}  |  (\theta  - \theta^\varepsilon)^\top y | + |  (\theta  - \theta^\varepsilon)^\top x | \\
    &\le 2 \max_{y \in \cX}  |  (\theta  - \theta^\varepsilon)^\top y | \\
    &\le  2 \| \theta  - \theta^\varepsilon\|_{\infty}  \max_{y \in \cX} (\indic^{\top} y) \\
    &\le  2 m \varepsilon
\end{align*}
Based on the above inequality, for any $x \in \cX^\star$ we have:
\begin{align*}
    \Delta_{x}(\theta^\varepsilon) &= \Delta_{x}(\theta) {\Delta_{x}(\theta^\varepsilon) \over \Delta_{x}(\theta)} 
    \le \Delta_{x}(\theta) {\Delta_{x}(\theta) + 2m\varepsilon \over \Delta_{x}(\theta)} \\
    &= \Delta_{x}(\theta) (1+ {2 m \varepsilon \over \Delta_{x}(\theta)}) 
    \le \Delta_{x}(\theta) (1+ {2 m \varepsilon \over \Delta_{\min}})
\end{align*}
Similarly, for any $x \in \cX^\star$ we have:
\begin{align*}
        \Delta_{x}(\theta) &= \Delta_{x}(\theta^\varepsilon) {\Delta_{x}(\theta) \over \Delta_{x}(\theta^\varepsilon) } 
        \le \Delta_{x}(\theta^\varepsilon) { \Delta_{x}(\theta^\varepsilon) + 2 m \varepsilon \over \Delta_{x}(\theta^\varepsilon) } \\
        &\le \Delta_{x}(\theta^\varepsilon) (1 + {2m \varepsilon \over \Delta_{x}(\theta^\varepsilon)})
        \le \Delta_{x}(\theta^\varepsilon) (1 + {4m \varepsilon \over \Delta_{\min}})
\end{align*}
where we used the fact that $$\Delta_{x}(\theta^\varepsilon) \ge \Delta_{x}(\theta) - 2m \varepsilon \ge {\Delta_{\min} \over 2}$$ 
since $\varepsilon \le {\Delta_{\min} \over 2m}$.

We now turn to the relationship between $P_{GL}$ and $P_{GL}^\varepsilon$. Consider $x \in \cX^\star$. Since $\alpha^{\star,\varepsilon}$ is an optimal solution to $P_{GL}^{\varepsilon}$, we must have 
\begin{align*}
    \sum_{i \in \cI} {x_i \over \sum_{y \in \cX} y_i \alpha^{\star,\varepsilon}_y} &\le ({\Delta_x(\theta^{\varepsilon})})^2 \le ({\Delta_x(\theta)})^2 (1 + {2m \varepsilon \over \Delta_{\min}})^2
\end{align*}
using our previous reasoning. We have  proven that  $\alpha^{\star,\varepsilon}(1 + {2m \varepsilon \over \Delta_{\min}})^2$ is a feasible solution to $P_{GL}$.

Using the same technique, since $\alpha^\star$ is an optimal solution to $P_{GL}$, we have that
$$
    \sum_{i \in \cI} {x_i \over \sum_{y \in \cX} y_i \alpha^{\star}_y} \le ({\Delta_x(\theta)})^2 \le ({\Delta_x(\theta^\varepsilon)})^2  (1 + {4m \varepsilon \over \Delta_{\min}})^2
$$
We have  proven that  $\alpha^{\star}(1 + {4m \varepsilon \over \Delta_{\min}})^2$ is a feasible solution to $P_{GL}^\varepsilon$.

We can now conclude:
\begin{align*}
    \sum_{x \in \cX^\star}  \alpha^{\star,\varepsilon} \Delta_{x}(\theta) 
    &\le (1 + {4m \varepsilon \over \Delta_{\min}}) \sum_{x \in \cX^\star} \alpha^{\star,\varepsilon} \Delta_{x}(\theta^\varepsilon) \\
    &\le (1 + {4m \varepsilon \over \Delta_{\min}})^3 \sum_{x \in \cX^\star} \alpha^{\star} \Delta_{x}(\theta^\varepsilon) \\
    &\le (1 + {4m \varepsilon \over \Delta_{\min}})^4 \sum_{x \in \cX^\star} \alpha^{\star} \Delta_{x}(\theta)
\end{align*}
where we successively used the inequality derived above, the fact that $\alpha^{\star}(1 + {4m \varepsilon \over \Delta_{\min}})^2$ is a feasible solution to $P_{GL}^\varepsilon$ and the inequality derived above again.

\section{Proofs}\label{section:proofs}
\subsection{Proof of Theorem~\ref{th:lower_bound}}
From \cite{combes2015}[Theorem 1], the result holds when $C(\theta)$ is the value of the following optimization problem:
\begin{align*}
    \underset{\alpha \in \RR^{|\cX|}_+}{\text{minimize   }} & \sum_{x \in \cX} \alpha_x \Delta_x  \\
    \text{subject to    } & \min_{\lambda \in B(\theta)} \left\{ \sum_{i=1}^d \sum_{x \in \cX}  \alpha_x x_i D(\theta_i,\lambda_i)\right\}  \ge 1
\end{align*}
where 
$$
    B(\theta) = \left\{\lambda \in \RR^d: \lambda^\top x^\star <  \max_{x \in \cX} \{ \lambda^\top x \} \text{ and } \theta_i = \lambda_i \;,\; \forall i \not\in \cI \right\}
$$
is the set of parameters $\lambda$ under which $x^\star$ is not the optimal decision, and such that $\lambda$ cannot be distinguished from $\theta$ when selecting only optimal decisions under $\theta$. $D(\theta_i,\lambda_i)$ is the Kullback Leibler divergence between the distribution of the rewards for $i$ with respective means $\theta_i$ and $\lambda_i$. Since rewards are Gaussian with variance ${1 \over 2}$, the divergence is given by $D(\theta_i,\lambda_i) = (\theta_i-\lambda_i)^2$. Furthermore, if $i \not\in \cI$, then $\theta_i = \lambda_i$, so that $D(\theta_i,\lambda_i) = 0$. Thus, the optimization problem simplifies to:
\begin{align*}
    \underset{\alpha \in \RR^{|\cX|}_+}{\text{minimize   }} & \sum_{x \in \cX} \alpha_x \Delta_x  \\
\text{subject to    } & \min_{\lambda \in B(\theta)} \left\{ \sum_{i \in \cI} \sum_{x \in \cX}  \alpha_x x_i (\theta_i-\lambda_i)^2 \right\}  \ge 1
\end{align*}
Decompose $B(\theta)$ according to the optimal decision and its value as follows:
\begin{align*}
    B(\theta) &= \cup_{v > 0} \cup_{x \ne x^\star} B_{x,v}(\theta) \\
    B_{x,v}(\theta) &= \{\lambda \in B(\theta): \lambda^\top x = \theta^\top x^\star + v\}
\end{align*}
Thus, the optimum solution of 
$$
    \min_{\lambda \in B_{x,v}(\theta)} \left\{ \sum_{i \in \cI} \sum_{x \in \cX} \alpha_x x_i (\theta_i-\lambda_i)^2\right\}
$$
is $\lambda \in \RR^d$ minimizing the quadratic function $\sum_{i \in \cI} \sum_{x \in \cX} \alpha_x x_i (\theta_i-\lambda_i)^2$ subject to the linear equality constraint $\lambda^\top x = \theta^\top x^\star + v$. Writing the Karush-Kuhn-Tucker conditions and solving, we can check that the minimum is:
$$
    \min_{\lambda \in B_{x,v}(\theta)} \left \{\sum_{i \in \cI} \sum_{x \in \cX} \alpha_x x_i (\theta_i-\lambda_i)^2 \right\} = {(\Delta_x + v)^2 \over \sum_{i \in \cI} x_i (\sum_{y \in \cX} y_i \alpha_y)^{-1} }
$$
The constraint
$$
\min_{\lambda \in B(\theta)} \left \{\sum_{i \in \cI} \sum_{x \in \cX} \alpha_x x_i (\theta_i-\lambda_i)^2 \right\} \ge 1
$$
is satisfied if and only if the above is greater than $1$ for all $x \in \cX$ and all $v > 0$, i.e:
$$
    \sum_{i \in \cI} {x_i \over \sum_{y \in \cX} y_i \alpha_y} \le \Delta_{x}^2 \;,\; \forall x \in \cX
$$
Therefore, the original optimization problem is, as claimed, the Graves-Lai optimization problem
\begin{align*}
    \hspace{3cm} \underset{\alpha \in \RR^{|\cX|}_+}{\text{minimize   }} & \sum_{x \in \cX} \alpha_x \Delta_x \hspace{4cm} \quad (P_{GL}) \\
    \hspace{3cm} \text{subject to    } & \sum_{i \in \cI} {x_i \over \sum_{y \in \cX} y_i \alpha_y} \le \Delta_x^2 \;,\; \forall x \in \cX
\end{align*}
This concludes the proof.

\subsection{Proof of Proposition~\ref{proposition:reduction}}

We start by stating the definition of $P_{GL}$, and notice that both the objective function and the constraints solely depend on $\sum_{x} \alpha_x$ and $\sum_{x} x \alpha_x$.
\begin{align*}
    \underset{\alpha \in \RR^{|\cX|}_+}{\text{minimize   }} & \sum_{x \in \cX} \alpha_x \Delta_x \quad \quad (P_{GL}) \\
    \text{subject to    } & \sum_{i \in \cI} {x_i \over \sum_{y \in \cX} y_i \alpha_y} \le \Delta_x^2 , x \in \cX
\end{align*}

%Computing $\cI$ can be done by solving $P_{LM}$ at most $d$ times.

Those variables live in the following set:
\begin{align*}
    \left\{ \left(  \sum_{x \in \cX} x \alpha_x,  \sum_{x \in \cX} \alpha_x \right) : \alpha\in \RR^{|\cX|}_+ \right\} &= \{ (w,v): {w \over v} \in \conv(\cX) , v \ge 0 \} \\
    &= \left\{ (w,v): A {w \over v} = b , {w \over v} \ge 0, v \ge 0 \right\} \\
    &= \left\{ (w,v): A w = v b , w \ge 0, v \ge 0 \right\}.
\end{align*}
where $\conv(\cX)$ denoted the convex hull of $\cX$, and we have used Assumption~\ref{assumption:convex_hull}.

If $b=0$, we simply have that $A w = 0$; otherwise, $A w = v b$. Therefore, $b^\top A w = v b^\top b = v \|b\|^2$ and $v = {b^\top A w \over \|b\|^2}$. This implies that $A w = v b$ if and only if
$$
  0 =  A w - v b =  A w - {b b^\top A  w \over \|b\|^2}  = M w
$$
by definition of $M$. Therefore,  
\begin{align*}
    \sum_{x \in \cX} \alpha_x \Delta_x &= \left(\theta^\top x^\star\right) \left(\sum_{x \in \cX} \alpha_x\right) -  \theta^\top \left(\sum_{x \in \cX} x \alpha_x\right) \\
    &= (\theta^\top x^\star) v + \theta^\top w \\
    &= \left(\theta^\top x^\star\right) {b^\top A w \over \|b\|^2} - \theta^\top w \\
    &= \left(\left(\theta^\top x^\star\right) {b^\top A \over \|b\|^2} -  \theta \right)^\top w \\
    &= q^\top w
\end{align*}
by definition of $q$. 

By Assumption~\ref{assumption:covering}, for any $i \in \cI$, there exists $x^i$ such that $x^i_i = 1$. As a consequence, for any feasible solution $w$,
$$
  {1 \over w_i} \le \sum_{j \in \cI} {x_j^i \over w_j} \le \Delta_{x^i}^2 \le (\theta^\top x^\star)^2 \le m^2 \| \theta \|_{\infty}^2 
$$
Thus, we can impose the additional constraint that $\min_{i \in \cI} w_i \ge \underline{w} \equiv (m \| \theta \|_{\infty})^{-2}$ for $i \in \cI$.

This yields the claimed reduced form:
\begin{align*}
    \underset{w \in \RR^{d}}{\text{minimize   }} &  q^\top w  \quad \quad (P_{GL}') \\
    \text{subject to } & \sum_{i \in \cI} {x_i \over w_i} \le \Delta_x^2 , x \in \cX , \;\;  M w = 0, w \ge 0,  \min_{i \in \cI} w_i \ge \underline{w},
\end{align*}
which concludes the proof of the proposition.

\subsection{Technical Lemma: Optimal Solution}

\begin{lemma}\label{lem:bound}
    Define $\alpha_x^{\star}$ an optimal solution to $P_{GL}$. Define $w^\star = \sum_{x \in \cX} x \alpha_x^\star$ the corresponding solution to $P_{GL}'$. 
    
    Then, its value is upper bounded by
    $$  
        q^\top w^\star =  \sum_{x \in \cX} \alpha_x^{\star} \Delta_{x} \le m d {\Delta_{\max} \over \Delta_{\min}^2}
    $$
    and the norm of the optimal solution is upper bounded by
    $$
        \|w^\star\| = \left\|\sum_{x \in \cX} x \alpha_x^\star \right\|  \le m^{3 \over 2} d {\Delta_{\max} \over \Delta_{\min}^2}
    $$
    Furthermore, if Assumption~\ref{assumption:integrality} holds, we have 
    $$
        q^\top w^\star \le m^2 d \| \theta \|_{\infty}
    $$
    and
    $$
        \|w^\star \| \le m^{5 \over 2} d \| \theta \|_{\infty}
    $$
\end{lemma}
Define $w^\star = \sum_{x \in \cX} x \alpha_x$. From Assumption~\ref{assumption:covering}, for each $i \in {\cal I}$, consider $x^{i} \in \cX$ such that $x^i_i = 1$. Consider 
$$
    w = \sum_{i=1}^{d} {m  \over \Delta_{\min}^2} x^{i} 
$$
This implies that for all $i=1,...,d$
$$
    w_i \ge {m \over \Delta_{\min}^2}
$$
and in turn for any $x$:
$$
      \sum_{i \in \cI} {x_i \over w_i} \le \Delta_{\min}^2 {1 \over m}   \sum_{i \in \cI} x_i 
      \le \Delta_{\min}^2 {1 \over m}  (\sum_{i=1}^d  x_i) 
      \le \Delta_{\min}^2 \le \Delta_{x}^2
$$
Hence, $w$ is a feasible solution, which implies that
$$
    q^\top w^\star \le q^\top w = \sum_{i=1}^d m {\Delta_{x^i} \over \Delta_{\min}^2}  \le m d {\Delta_{\max} \over \Delta_{\min}^2}
$$
Now, by definition the optimal solution can be expressed as:
$$
    w^\star = \sum_{x \in \cX} x \alpha_x
$$
First, notice that
$$
q^\top w^\star = \sum_{x \in \cX} \alpha_x \Delta_x \ge \Delta_{\min} \sum_{x \in \cX} \alpha_x
$$
Consequently,
$$
    \sum_{x \in \cX} \alpha_x \le {q^\top w^\star \over \Delta_{\min}}
$$
Using the triangle inequality:
$$
     \left\|\sum_{x \in \cX} x \alpha_x \right\| \le \sum_{x \in \cX} \alpha_x \|x\| \le \sqrt{m} \sum_{x \in \cX} \alpha_x 
     \le  m { q^\top w^\star \over \Delta_{\min}}
     \le m^2 d {\Delta_{\max} \over \Delta_{\min}^2}
$$
This proves the first result. 

If Assumption~\ref{assumption:integrality} holds as well, we have 
$$
    1 \le \Delta_{\min} \le \Delta_{\max} \le m \| \theta \|_{\infty}
$$
which proves the second result.

\subsection{Proof of Proposition~\ref{prop:compute_xt}}

From Assumption~\ref{assumption:integrality}, $\theta$ has positive integer components: for any $x \in \cX$, we have $\theta^\top x \in \{0,...,m \| \theta \|_{\infty}\}$. In turn, this implies that $\Delta_{x} \in \{0,...,m \| \theta \|_{\infty}\}$ for all $x \in \cX$. Now, we use Assumption~\ref{assumption:polynomial_BLM} or \ref{assumption:polynomial_BLM_approximate} to compute (in polynomial time), for $s \in \{0,..., m \| \theta \|_{\infty}\}$, an $\varepsilon$-approximate solution to $P_{BLM}$ denoted by $X^{t,s} \in \cX$ with
$$
    \sum_{i \in \cI} {X^{t,s}_i \over w_i^t}  \ge \varepsilon \left( \max_{x \in \cX:  \Delta_{x} \le s} \sum_{i \in \cI} {X^{t,s}_i \over w_i^t}\right) \text{ and } \Delta_{x} \le s
$$
and one may readily check that
\begin{align*}
    x^{t} = X^{t,s^t} \text{ with } s^t \in \arg\max_{s \in \{0,..., m \| \theta \|_{\infty}\}} \left\{  \sum_{i \in \cI} {X^{t,s}_i \over w_i^t} -  s^2 \right\}
\end{align*}
satisfies $\max_{x \in \cX} h_x(w^t) \le h_{x^t}(\varepsilon w^t)$. In summary, $x^{t}$ can be computed in time $O(\poly(d,\delta,\|\theta\|_{\infty}))$.

\subsection{Technical Lemma: Gradient Descent}

We first state a technical lemma due to \cite{shai2014}[Lemma 14.1].

\begin{lemma}\label{lem:shai}
	Consider ${\cal M}$ a convex set, $\eta > 0$, $\hat w$ and $g^1,...,g^T$ arbitrary vectors, $w^n$ a sequence defined as
	$$
		w^{t+1} = \Pi_{{\cal M}}\{ w^t - \eta g^t \}
	$$
	with $\Pi_{{\cal M}}$ the orthogonal projection onto $\cal M$.

	Then we have:
	$$
		 \sum_{t=1}^T \langle w^t - \hat{w}, g^t \rangle \le {\| \hat{w} \|^2 \over 2 \eta} + {\eta \over 2} \sum_{t=1}^T \| v^t \|^2
	$$
\end{lemma}

\cite{shai2014}[Lemma 14.1] first states the lemma without the projection step, and afterwards argue that their proof still holds when a projection step is added, which corresponds to Lemma~\ref{lem:shai}.

\subsection{Proof of Proposition~\ref{prop:gradient_descent}}

The procedure~\ref{algo_complete} runs in time $O(\poly(d,\delta,\|\theta\|_{\infty}))$. Indeed, the number of iterations is $T = O(\poly(d,\delta,\|\theta\|_{\infty}))$, and each iteration takes time $O(\poly(d,\delta,\|\theta\|_{\infty}))$.

Define the error:
$$
	E =  q^\top \bar{w} - q^\top (w^\star/\varepsilon) + \lambda \max_{x \in \cX}(h_x(\bar{w}))^+
$$
Using Jensen's inequality, since $w \mapsto \max_{x \in \cX} (h_x(w))^+ $ is convex,
$$
	E \le {1 \over T} \sum_{t=1}^T (q^\top w^t) - q^\top(t^\star/ \varepsilon) + \lambda {1 \over T} \sum_{t=1}^T \max_{x \in \cX} (h_x(w^t))^+
$$
We use the following notation for the dot product:
$$
q^\top w^t - q^\top(w^\star/ \varepsilon) = \langle w^t - w^\star/\varepsilon , q \rangle
$$
By definition of $x^t$:
$$
	\max_{x \in \cX} (h_x(w^t))^+ \le (h_{x^t}(\varepsilon w^t))^+
$$
Using the fact that $w \mapsto (h_{x^t}(\varepsilon w))^+$ is a convex function and one of its subgradients is
$$
\varepsilon \nabla h_{x^t}(\varepsilon w) \indic( h_{x^t}(\varepsilon w) > 0)
$$
we get, by definition of a subgradient:
$$
	(h_{x^t}(\varepsilon w))^+ - (h_{x^t}(\varepsilon (w^\star/\varepsilon)))^+ \le  \langle w^t - w^\star/\varepsilon , \varepsilon \nabla h_{x^t}(\varepsilon w^t) \indic( h_{x^t}(\varepsilon w^t) > 0) \rangle
$$
We have that $h_{x}(w^\star) < 0$ for all $x$ by definition of $w^\star$, so that $(h_{x^t}(\varepsilon (w^\star/\varepsilon)))^+ = 0$ and replacing above we get:
$$
	\max_{x \in \cX} (h_x(w^t))^+ \le  \langle w^t - w^\star/\varepsilon , \varepsilon \nabla h_{x^t}(\varepsilon w^t) \indic( \nabla h_{x^t}(\varepsilon w^t) > 0) \rangle
$$
Thus:
$$
	E \le {1 \over T} \sum_{t=1}^T   \langle w^t - w^\star/\varepsilon  ,  q + \lambda \varepsilon \nabla h_{x^t}(\varepsilon w^t) \indic( h_{x^t}(\varepsilon w^t) > 0) \rangle
$$
Using the fact that:
\begin{align*}
	w^{t+1} &= \Pi_{\cM} \Big\{ w^{t} - \eta g^t \Big\}. \\
	g^t &=  q + \lambda \varepsilon \nabla h_{x^t}(\varepsilon w^t) \indic( h_{x^t}(\varepsilon w^t) > 0) \\
	\bar{w} &= {1 \over T} \sum_{t=1}^{T} w_t. 
\end{align*}	
Lemma~\ref{lem:shai} yields:
$$
E \le {1 \over 2 T} \left(  {\|w^0 - w^{\star}/\varepsilon  \|^2 \over \eta} + \eta \sum_{t=1}^T \| q + \lambda \varepsilon \nabla h_{x^t}(\varepsilon w^t) \indic( h_{x^t}(\varepsilon w^t) > 0)  \|^2  \right)
$$
so that 
$$
E \le {1 \over 2 T} \left( {\|w^0 - w^{\star}/\varepsilon \|^2 \over \eta} + \eta \sum_{t=1}^T ( \| q \|^2 + (\lambda \varepsilon)^2 \| \nabla h_{x^t}(\varepsilon w^t)  \|^2  ) \right)
$$
We may upper bound each term in the expression above as follows.

Since
$$
	\nabla h_{x^t}(\varepsilon w^t) = -\left({x_1^t \over (\varepsilon w_1)^2},..., {x_d^t \over (\varepsilon w_d^t)^2} \right),
$$
the gradient term is upper bounded as
$$
	\|\nabla h_{x^t}(\varepsilon w^t) \|^2 =  \sum_{i\in \cI} {x_i^t \over (\varepsilon w_i^t)^4}  \le {d \over (\varepsilon  \underline{w})^4} = \varepsilon^{-4} d m^8 \|\theta\|_{\infty}^8.
$$
where we used the fact that $w^t \in \cM$, which implies $w^t \ge \underline{w}$.

Since $w^0 = (\underline{w},...,\underline{w})$ and $w^\star/\varepsilon \ge \underline{w}$, we get:
$$
    \|w^0 - w^{\star}/\varepsilon  \| \le \varepsilon^{-1} \| w^{\star} \| 
    \le \varepsilon^{-1} m^{5 \over 2} d \| \theta \|_{\infty} 
$$
using Lemma~\ref{lem:bound}.

Replacing, we get the upper bound:
$$
E \le {1 \over 2 T} \left(  {\varepsilon^{-2} m^{5} d^2 \| \theta \|_{\infty}^2 \over \eta} + \eta T \left( \| q \|^2 + \lambda^2 \varepsilon^{-2} d m^8 \|\theta\|_{\infty}^8 \right) \right)
$$
setting $\eta$ to equalize both terms
$$
	\eta^2 = { \varepsilon^{-2} m^{5} d^2 \| \theta \|_{\infty}^2 \over T ( \| q \|^2 + \lambda^2 \varepsilon^{-2} d m^8 \|\theta\|_{\infty}^8)}  
$$
so that the optimization error has the following upper bound:
$$
	E \le {\varepsilon^{-2} m^{5} d^2 \| \theta \|_{\infty}^2 \over \eta T}
	= {1 \over \sqrt{T}} \varepsilon^{-1} m^{5/2} d \| \theta \|_{\infty} \sqrt{ \| q \|^2 + \lambda^2 \varepsilon^{-2} d m^8 \|\theta\|_{\infty}^8}
$$
Recall the definitions:
$$
    \delta_{2} = {\delta \varepsilon \over m d \| \theta \|_{\infty}}
$$
and
$$
    \delta_{1} = {\delta \over 2 (1+\delta_2)}
$$
Now, setting
$$
T = {1 \over \delta_1^2} \varepsilon^{-2} m^{5} d^2 \| \theta \|_{\infty}^2 \left( \| q \|^2 + \lambda^2 \varepsilon^{-2} d m^8 \|\theta\|_{\infty}^8\right)
$$
we get that $E \le \delta_1$. Replacing $E$ by its definition, this proves that:
$$
        E = q^\top \bar{w} - q^\top (w^\star/\varepsilon) + \lambda \max_{x \in \cX}(h_x(\bar{w}))^+ \le \delta_1
$$
This allows to upper bound the constraints violation:
$$
    \lambda \max_{x \in \cX}(h_x(\bar{w}))^+ \le {\delta_1 - q^\top \bar{w} + q^\top (w^\star/\varepsilon) \over \lambda}
    \le {\delta_1 + {m^2 d \| \theta \|_{\infty} } \over \lambda}
$$
using the fact that $q^\top \bar{w} \ge 0$ and $q^\top w^\star \le m d \| \theta \|_{\infty}$ from Lemma~\ref{lem:bound}. Setting 
$$
\lambda = {1 \over \delta_2} (\delta_1 + {m^2 d \| \theta \|_{\infty} })
$$
this proves that 
$$
     \max_{x \in \cX}(h_x(\bar{w}))^+ \le \delta_2
$$
Define $\bar{w}' = (1 + \delta_2) \bar{w}$. Since, for all $x$,
$$
    \sum_{i \in \cI} {x_i \over \bar{w}_i} \le \Delta_x^2 + \delta_2
$$
we have that
$$
    \sum_{i \in \cI} {x_i \over \bar{w}_i'} \le {\Delta_x^2 + \delta_2 \over 1 + \delta_2} 
    = \Delta_x^2 { {1 \over \Delta_x^2} + \delta_2 \over 1 + \delta_2 } \le \Delta_{x}^2
$$
using the fact that $\Delta_x^2 \ge 1$ from Assumption~\ref{assumption:integrality}. Hence, $\max_{x \in \cX}(h_x(\bar{w}'))^+ = 0$, which means that $\bar{w}'$ is a feasible solution.

Finally:
$$
    q^\top \bar{w} - q^\top (w^\star/\varepsilon)  \le q^\top \bar{w} - q^\top (w^\star/\varepsilon) + \lambda \max_{x \in \cX}(h_x(\bar{w}))^+ \le \delta_1
$$
so that
$$
    q^\top \bar{w}' - q^\top (w^\star/\varepsilon) \le (1+\delta_2)  \delta_1 + \delta_2 q^\top (w^\star/\varepsilon)
    \le (1+\delta_2)  \delta_1 + \delta_2 m^2 d \| \theta \|_{\infty}  /\varepsilon = {\delta \over 2} + {\delta \over 2} = \delta.
$$
Putting it all together, we have proven that $\bar{w}'$ is a feasible solution which verifies 
$$
    q^\top \bar{w}' - q^\top (w^\star/\varepsilon) \le \delta
$$
This concludes the proof.

\subsection{Proof of Proposition~\ref{prop:caratheodory}}

We prove the result using recursion. Assume that $\bar{w}^k$ can be written as a linear combination with positive coefficients of elements of $\cX$. Define $\cJ^k  = \{i=1,...,d: w_i > 0\}$ the set of its non-null entries. We have $x^k \in \arg\min_{x \in \cX} \sum_{i \not\in \cJ} x_i$.

Since, by assumption, $\bar{w}^k$ can be written as a linear combination with positive coefficients of elements of $\cX$, there exists $x$ such that $\sum_{i \not\in \cJ^k} x_i = 0$. Therefore, by definition of $x^k$, we must have $0 \le \sum_{i \not\in \cJ^k} x_i^k  \le \sum_{i \not\in \cJ^k} x_i = 0$, so that $\sum_{i \not\in \cJ} x_i^k = 0$. We then write 
$$
     \bar{w}^{k+1} =  \bar{w}^k - x^k \alpha_{x^k}
$$
Now, we have that $\bar{w}^{k+1} \ge 0$ since $\bar{w}^{k+1}_i = \bar{w}^{k}_i$ if $i \not\in \cJ^k$ and $\bar{w}^{k+1}_i = \bar{w}^{k}_i - \min_{i \in \cJ^k} \bar{w}^{k}_i \ge 0$ if $i \in \cJ^k$. Furthermore, we have that 
$$
    M \bar{w}^{k+1} = \alpha_{x^k}  M x^k  +  M \bar{w}^{k} = M \bar{w}^{k} = 0
$$
since $x^k \in \conv(\cX)$, which implies $M x^k = 0$. Therefore, $\bar{w}^{k+1}$ can be written as a linear combination with positive coefficients of elements of $\cX$. Also, $|\cJ^{k+1}| \le \max(0,|\cJ^k| - 1)$: indeed, we have that $\cJ^{k+1} \subset \cJ^k$ and, if $i_k \in \arg\min_{i \in \cJ^k} w^k_i$, we have that $w_{i_k}^k > 0$ and $w_{i_k}^{k+1} = 0$ by construction.

Since $w^1 = w^\star$ can be written as a linear combination with positive coefficients of elements of $\cX$, the above argument shows that $w^k$ can be written as a linear combination with positive coefficients of elements of $\cX$ for all $k$, and that $w^{d+1} = 0$. This implies that the procedure does terminate after at most $d$ iterations and
$$
    w^\star = \sum_{k=1}^d x^k \alpha_{x^k}
$$
with $x^1,...,x^d$ in $\cX$ and $\alpha_{x^1},...,\alpha_{x^d}$ positive numbers.

For each iteration, it is noted that $x^k$ can be computed by linear maximization over $\cX$, which is feasible in time $O(\poly(d))$ from Assumption~\ref{assumption:polynomial_LM}. Since the above procedure terminates after at most $d$ iterations, it takes $O(\poly(d))$ time.

\subsection{Projection Step}\label{ref:projection_step}

The projection of $w^{t} - \eta g^t$ on $\cM$ involves solving the following optimization problem:

\begin{align*}
    \underset{w \in \RR^{d}}{\text{minimize   }} &   \|w - w^{t} - \eta g^t \|^2  \quad \quad  \\
    \text{subject to } & M w = 0, w \ge 0, \min_{i \in \cI} w_i \ge \underline{w}.
\end{align*}

There are two possible cases: \textit{(i)} in some cases, this projection may be computed exactly, \textit{(ii)} the projection may be computed using an interior point method using a logarithmic barrier function and a Newton step \cite{boyd2004} Chapter 11.2. By definition, $w_t \in {\cal M}$, so that $w_t$ can be used as an initial feasible point to compute the projection. We also mention that there exists even more efficient algorithms for specific combinatorial sets, for instance for the matching polytope~\cite{wang2010learning}.

\end{document}